# Multi-cropping Contrastive Learning and Domain Consistency for Unsupervised Image-to-Image Translation

Chen Zhao, Wei-Ling Cai, Zheng Yuan and Cheng-Wei Hu, *Member, IEEE*

*Abstract*—Recently, unsupervised image-to-image translation methods based on contrastive learning have achieved state-of-the-art results in many tasks. However, in the previous works, the negatives are sampled from the input image itself, which inspires us to design a data augmentation method to improve the quality of the selected negatives. Moreover, the previous methods only preserve the content consistency via patch-wise contrastive learning in the embedding space, which ignores the domain consistency between the generated images and the real images of the target domain. In this paper, we propose a novel unsupervised image-to-image translation framework based on multi-cropping contrastive learning and domain consistency, called MCDUT. Specifically, we obtain the multi-cropping views via the center-cropping and the random-cropping with the aim of further generating the high-quality negative examples. To constrain the embeddings in the deep feature space, we formulate a new domain consistency loss, which encourages the generated images to be close to the real images in the embedding space of the same domain. Furthermore, we present a dual coordinate attention network by embedding positional information into the channel, which called DCA. We employ the DCA network in the design of generator, which makes the generator capture the horizontal and vertical global information of dependency. In many image-to-image translation tasks, our method achieves state-of-the-art results, and the advantages of our method have been proven through extensive comparison experiments and ablation research.

## I. Introduction

Image-to-image translation plays a vital role in many applications [1][2][3], such as super resolution[1][2], style transfer[3][4], deraining[5], image restoration [6], image colorization[7], dehazing and denoising[8][9]. This task aims at mapping an input image from one domain (source domain) into another (target domain) while preserving its original content, such as cat to dog, horse to zebra, apple to orange, low resolution image to high resolution image, photography to painting, etc.

Manuscript is submitted on January 17, 2023. This work was supported by the National Natural Science Foundation of China (Grant No. 62276138). (Corresponding Author: Wei-Ling Cai)

Chen Zhao, Wei-Ling Cai, Zheng Yuan and Cheng-Wei Hu are with the Department of Computer Science & Technology, Nanjing Normal University, Nanjing 210023, China (e-mail: 2518628273@qq.com, 24577075@qq.com, 1455341238@qq.com, 2453347006@qq.com).

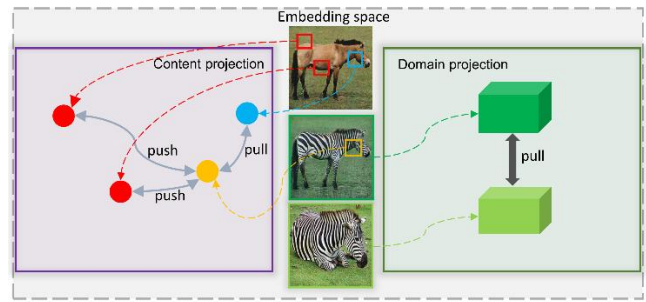

**Fig.1.** The motivation of the domain consistency. The previous works retain the content consistency of the generated images via patch-wise contrastive learning [12] in the embedding space. However, they ignore the domain consistency between the generated image and the real image of the target domain.

Since paired training data is difficult to obtain, current methods are mostly based on unpaired settings. In unpaired image-to-image translation, there exist many possible mappings between the two domains, which makes the adversarial loss[10] unconstrained. Therefore, in order to limit the adversarial loss, the cycle consistency loss[11] is used to constrain the training of the model and maintain the content consistency between the original image and the generated image. However, the assumption of cycle consistency that two domains can be mapped in both directions is too strict to obtain sufficient context[12].

Due to the great success of contrastive learning in the field of self-supervised learning[13][14], Contrastive learning for Unpaired image-to-image Translation[12] (CUT) introduced contrastive learning into image-to-image translation for the first time. By incorporating contrastive learning between the source and target domains, CUT[12] aims to use auxiliary knowledge by the generator to supplement the cycle consistency[15]. By maximizing mutual information between input and output images in the embedding space, CUT[12] achieved SOTA results in unpaired image-to-image translation.

Given a query in the generated image, one corresponding positive is sampled from the input image. Meanwhile, many negatives are also drawn from non-local patches in the input images. Although this flexible method improves the quality of image translation, viewing all non-local patches as negative samples leads to an obvious issue, that is, the unbalanced distribution of such negatives may increase the difficulty of learning[16]. Therefore, NEGCUT[16] pointed out that the negatives play an important role in the contrastive learning and proposed a hard negatives generation method. However, this method needs to train an additional generator, which greatly increases the training difficulty. DCLGAN[17] proposed a method based on contrastive learning and dual learning settings (using two generators and discriminators) to infer effective mappings between unpaired data. The dual settings are helpful to the stability of training. Although improving the quality, an additional generator and discriminator need to be trained, which increase the training costs.

In previous studies, the query for contrastive learning is randomly selected from the generated images, which is an obvious problem since some locations contain less information of the source domain[18]. QS-Attn[18] solved the problem by deliberately choosing important anchors for contrastive learning. A query selection attention module is designed, which compares the feature distance of the source domain and gives the attention matrix of the probability distribution of each row.

In the previous unsupervised image-to-image translation methods based on contrastive learning, no matter CUT[12], DCLGAN[17], NEGCUT[16], QS-Attn[18], the negatives are from the input image itself, which inspires us to design a data augmentation method to improve the quality of the selected negatives. Moreover, the previous methods only preserve content consistency via maximizing mutual information between the input and the generated image in the embedding space, ignore domain consistency between the generated image and the real image of target domain. **Fig.1** shows our motivation of the domain consistency.

In this paper, we propose an unpaired image-to-image translation framework based on multi-cropping contrastive learning and domain consistency, called MCDUT. At first, we propose a negative sample generation method based on multi-crop views to improve the quality of negative samples. This method utilizes center-cropping and random-cropping methods to randomly generate the multiple views from the input image. The negatives are selected from the generated multi-cropping views. The center-cropping method, which focuses mainly on the foreground, allows us to obtain the key information of the input images. The random-cropping views try to focus on the background, which can complement the limitations of center-cropping and make the negatives more diverse. To constrain the domain consistency between the generated image and the real images, we propose a domain consistency loss in the deep feature space to improve the quality of the generated images. Furthermore, we present a dual coordinate attention (DCA) network by embedding positional information into the channel. We employ the DCA network in the design of generator to aggregate information from a channel to apply to its content, which makes the generator pay more attention to channels with greater weight. Compared with current attention module [19][20], DCA mainly incorporates global average pooling (GAP) coordinate information and global max pooling (GMP) coordinate information to the channel, which can make channel receive global mutual information of positions. In this way, each location can capture the horizontal and vertical global information of dependency. To evaluate the effectiveness of MCDUT, the comprehensive experiments are conducted among several methods[11][12][16][17][18] and the results

prove that MCDUT achieves SOTA results in multiple tasks. Its contributions are as follows:

- We propose the multi-cropping contrastive learning to maximize the mutual information between input and output images. We can obtain multi-cropping views from the input image via the multi-cropping method, which can improve the quality of the selected negatives.
- To constrain the embeddings in the deep feature space, we present a loss function that focuses on the domain of the images, namely, domain consistency loss, which constrains features from the same domain.
- We present a new dual coordinate attention (DCA) network, which adopts GAP and GMP coordinate information into the channel. In this way, we can increase the receptive field of network and enhance the feature representation abilities of image-to-image translation.

## II. RELATED WORKS

### A. Image-to-image translation

In many image applications, GANs [10] have obtained great success, especially in image-to-image translation and the key idea is adversarial loss [10]. Image-to-image translation aims to learn a map from a source domain into a target domain [11][21], which can be categorized into two groups: a paired setting[22][24] (supervised) and an unpaired setting (unsupervised)[11][25]. Paired setting means that each image from the source domain has a corresponding label from the target domain. Pix2Pix[22] first proposed a general framework, supporting multiple image-to-image translation tasks. It was extended to high-resolution in Pix2PixHD[23], which can be considered as classical GAN. In order to further improve the quality of the generated images, SPADE[24] introduced the spatially-adaptive normalization layer.

However, it is difficult to obtain paired training data, as a result, current methods[10] are usually based on unpaired settings, which are mainly developed based on two assumptions: a shared latent space[24] and a cycle-consistency assumption[11]. For example, UNIT [25], Dual-GAN[26] and MUNIT[27] train cross-domain GANs with cycle-consistency loss[11]. However, the assumption of cycle consistency which two domains can be mapped in both directions is too strict to obtain sufficient context. To alleviate this issue, many methods[28][29] have tried to break the cycle consistency[11]. DistanceGAN[28] and GCGAN[29] allowed one-way translation. DistanceGAN[28] proposed a distance constraint that allows unsupervised domain mapping to be one-sided. GC-GAN[29] enforced geometry consistency as a constraint for unsupervised domain mapping. CUT[12] first tried to introduce contrastive learning into image-to-image translation, which significantly improved the quality of translation. F-LseSim[30] extended CUT[12] by computing the self-similarity within a local region, and imposed contrastive loss on it. However, it relies on VGG features to measure similarity, which reduces training efficiency. DCLGAN[16] proposed a method based on contrastive learning and dual learning settings (using two generators and discriminators) to infer effective mappings between unpaired data. The dual settings[25] are helpful to the stability of training. However, negatives are still from the input image itself, and these methods suffer from a number of negatives with low quality.

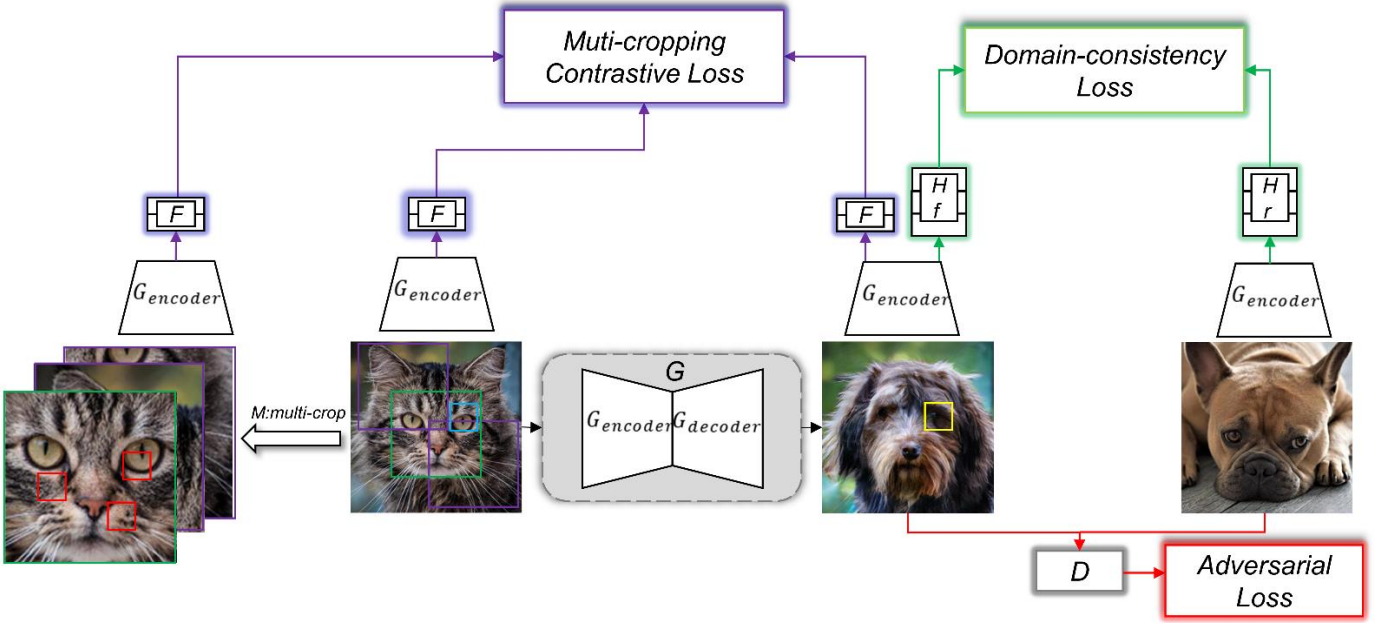

**Fig.2** Overall architecture of MCDUT. The objective function of MDCUT consists of four parts: adversarial loss[9] (red arrow), multi-cropping contrastive loss (purple arrow), domain consistency loss (green arrow), and identity loss[10]. The details of our objective are described below.

*B. Contrastive learning*

Recent studies of the self-supervised learning [31][32] show its strong ability to represent an image without labels, particularly with the help of the contrastive loss [13][14]. Its idea is to perform the instance level discrimination and learn the feature embedding, by pulling the features from the same image together and pushing those from different ones away[13][14]. Recently, it has been investigated as a pre-training technique[33][34], providing the initial model or the latent embedding for the down-stream task. PatchNCE[12] proposed patch-based contrastive learning, which uses a noise-contrastive estimation framework in unpaired image to image translation tasks by learning the correspondence between the patches of the input image and the corresponding generated image patches. Excellent results were achieved and the recent methods also obtained better performance by utilizing the idea of patch-wise contrastive learning. NEGCUT[16][15] proposed a method of hard negatives generation. QS-Attn[17] proposed a query-selected attention module by deliberately choosing important anchors for comparative learning. In parallel to these various designed methods, we mainly explore negatives selection strategies by introducing the idea of multi-cropping contrastive learning, which consistently shows better results.

*C. Attention mechanism*

Humans can naturally and efficiently detect salient regions in complex scenes. Motivated by this observation, attention mechanisms were introduced into computer vision with the aim of imitating this aspect of the human visual system. This attention mechanism can be viewed as a dynamic weight adjustment process based on input image features[19]. Attention mechanisms have been used with great success in many vision tasks, including image classification, target detection[38][39].

SENet[19] introduces channel attention mechanism, which obtains interdependencies among channels via squeezes each 2D feature map. CBAM[36] further promotes this idea by introducing spatial information. Later works[40][41][42] extended this idea by capturing different types of spatial information or designing different attention blocks. Self-Attention GAN[43] combine GAN and the self-attention mechanism. Many works[44] [48] added spatial-wise attention in GAN. CBAM-GAN[49] introduce CBAM module in GANs to extract more detail features. In order to improve quality of generated images, we try to introduce attention mechanism into image-to-image translation to enhance feature representation abilities.

III. METHOD

*A. Overall architecture*

Suppose there are two domains $\mathcal{X} \subset \mathbb{R}^{H \times W \times 3}$ and $\mathcal{Y} \subset \mathbb{R}^{H \times W \times 3}$, given unpaired datasets $X = \{x \in \mathcal{X}\}$ and $Y = \{y \in \mathcal{Y}\}$, our goal is to learn a mapping from input domain $\mathcal{X}$ to output domain $\mathcal{Y}$ : $G$: $X \rightarrow Y$. **Fig.2** shows the overall architecture of MCDUT. MCDUT consists of a generator $G$ and a discriminator $D$. $G$ creates a mapping from $\mathcal{X}$ domain to $\mathcal{Y}$ domain. $D$ is used to judge whether the input is real image from $\mathcal{Y}$ domain. We divide the generator into two parts: the encoder $G_{encoder}$ and the decoder $G_{decoder}$. In $G_{encoder}$, we add DCA network to better guide image generation. $M$ represents multi-cropping method, and multi-cropping views are obtained from input image by $M$. We extract features of images by $G_{encoder}$ and send them to the content projection network $F$, a two-layer MLP. Such a MLP learns to project the extracted features from $G_{encoder}$ to a stack of features. Additionally, we introduce two domain projection networks ($Hf$ and $Hr$) to extract the style information, which forms a domain consistency loss. $Hf$ and $Hr$ are used to extract the style information from the generated images and real $\mathcal{Y}$ domain image respectively.

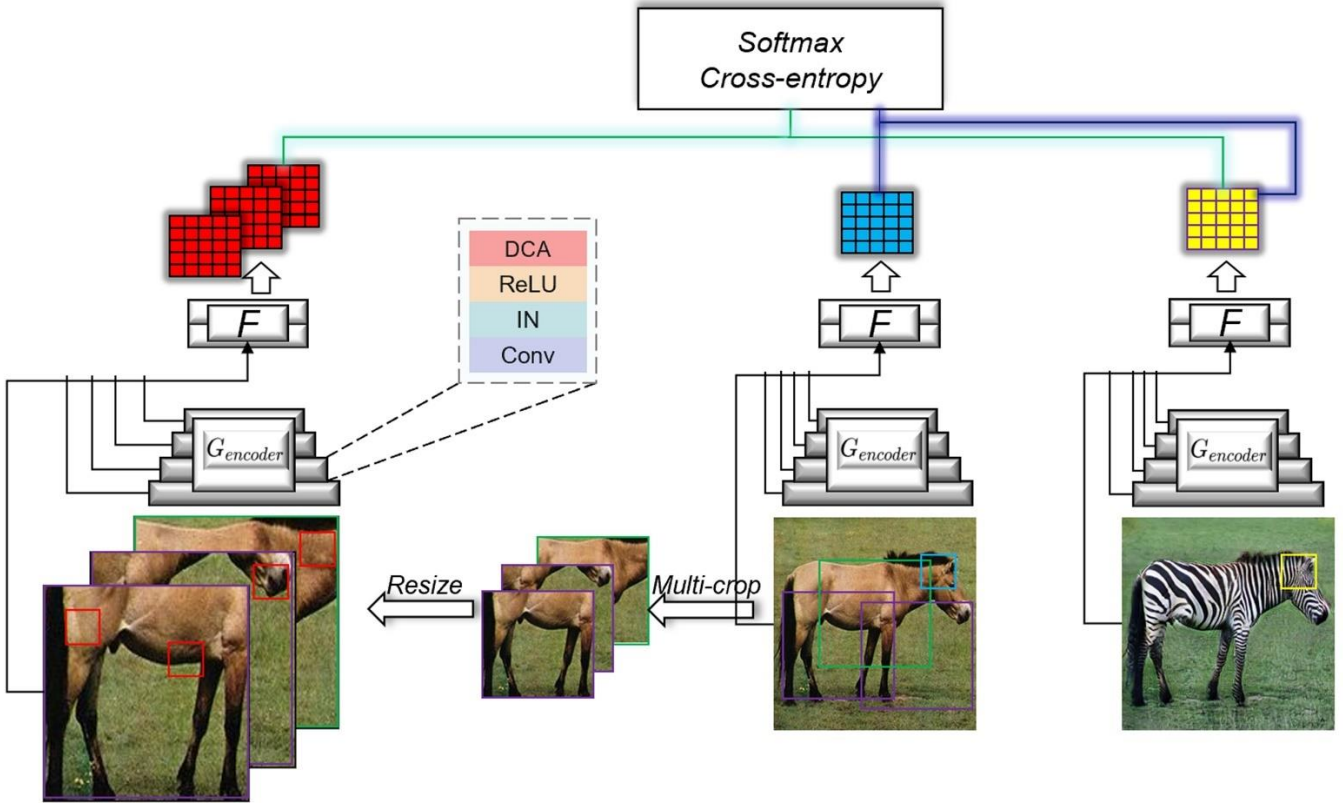

**Fig.3.** Multi-cropping contrastive learning. The yellow box refers to query, the blue box refers to positive example, and the red box refers to negative examples. We obtain the green box and purple box via center-crop method and random-crop method, respectively, which consist of our multi-cropping views. Moreover, we introduce our proposed dual coordinate attention (DCA) network into the encoder of generator.

*B. Multi-cropping contrastive learning*

We expect each position of the generated image to retain as much information as possible about the corresponding position of the input image. For example, the head of the generated zebra image should be very similar to the head of the horse in the input image, but not similar to other parts of the horse in the input image. We use a contrastive learning framework to maximize mutual information between patches between the input image and the generated image. The key idea of contrastive learning is to train an encoder which pulls the query and matched key (positives) as close as possible, and pushes query and non-match keys (negatives) as far as possible.

We represent the query, positive and $N$ negatives as $\boldsymbol{v}, \boldsymbol{v}^+ \in \mathbb{R}^K$ and $\boldsymbol{v}^- \in \mathbb{R}^{N \times K}$, respectively. $v_n^- \in \mathbb{R}^K$ is the $n$-th negative that is mapped to the $K$ dimension space. To set up a ($N+1$) classification problem, we use L2 normalization to distribute these vectors onto a unit sphere. Then we calculate the probability of positives and negatives being selected. The contrastive learning can be expressed as a classification problem, which is calculated by cross entropy loss. The formula can be expressed as:

$$\ell(\boldsymbol{v}, \boldsymbol{v}^+, \boldsymbol{v}^-) = -\log\left[\frac{\exp(\boldsymbol{v} \cdot \boldsymbol{v}^+ / \tau)}{\exp(\boldsymbol{v} \cdot \boldsymbol{v}^+ / \tau) + \sum_{n=1}^{N} \exp(\boldsymbol{v} \cdot \boldsymbol{v}_n^- / \tau)}\right], \quad (1)$$

where $\tau$ indicates a temperature parameter used to measure the distance between query and other samples. The default value is 0.07.

Multi-cropping contrastive learning is shown in **Fig.3.** We hope that each patch (yellow box) in the generated image can be closely connected with the corresponding patch (blue box) in the input image.

We utilize $G_{encoder}$ to extract features from the generated image, select the L layers of interest from $G_{encoder}$, and sent it to F to get the features we need. The resulted features can be denoted by $\{\hat{\boldsymbol{z}}_l\}_L = \{F_l(G_{encoder}^l(G(\mathbf{x})))\}_L$, which $G_{encoder}^l$ represents $l$-th layer we choose. We select patches in each selected layer, $s \in \{1, \ldots, S_l\}$ ($S_l$ represent the number of patches selected in each layer). In the same way, the corresponding patches of L layer are obtained from the input image, $\{\boldsymbol{z}_l\}_L = \{F_l(G_{encoder}^l(\mathbf{x}))\}_L$. We take the patches obtained from the generated image as a query (yellow square) and the corresponding patches obtained from the input image as a positives (blue square). Both of them participate in the part of cross entropy loss[50] (indicated by the purple line connecting the blue square and yellow square).

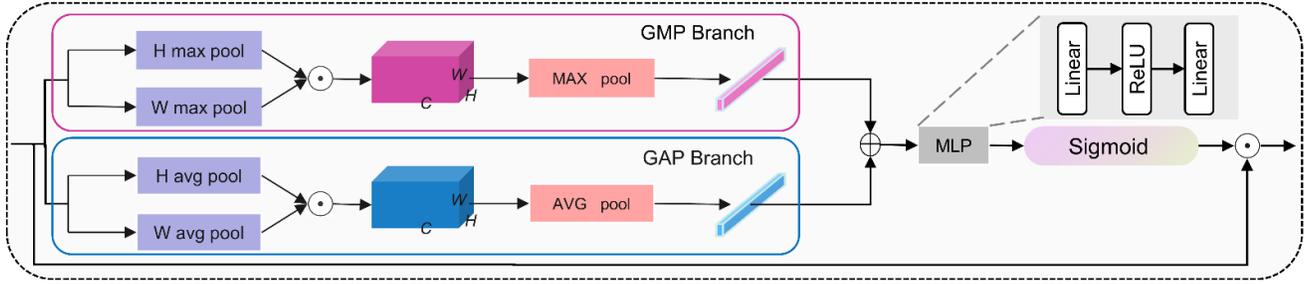

**Fig.4.** The detail framework of dual coordinate attention (DCA) network. DCA mainly consists of GAP Branch and GMP Branch. Here, "GAP" and "GMP" refer to the global average pooling and global max pooling, respectively. 'H Avg Pool' and 'W Avg Pool' refer to 1D horizontal global average pooling and 1D vertical global average pooling, respectively. 'H Max Pool' and 'W Max Pool' refer to 1D horizontal global max pooling and 1D vertical global max pooling, respectively.

We use a novel method to obtain negatives. We use the center-cropping method (the green box) to segment the input image. The content of the obtained center image is mainly concentrated in the foreground, including the main information of the image. Then we use the random-cropping method (the purple box) to supplement the limitations of center-cropping method and make negatives richer. We call this method as multi-cropping and the obtained multi-cropping views are shown in **Fig. 3**. The information of multi-cropping views is mainly concentrated in the foreground. Compared with the original input image, we can get more useful negatives. In order to distinguish from the positives of the input image and make the important negatives more, we resize after multi-cropping to obtain a multi-cropping images of the same size as the original input image. The negatives is selected in **Fig. 3** (the red box).

We employ $G_{encoder}$ and $F$ to obtain features from the multi-cropping images. Each cropping image can get the corresponding features (red squares), where each feature represents a patch in the cropping image, $\{m_l\}_L = \{F_l(G^l_{encoder}(M(x)))\}_L$.

We randomly select $N$ patches as negatives (pink squares). Negatives and the query (yellow square) obtained in the generated image participate in the part of cross entropy loss[50] (indicated by the green line connecting the pink square and yellow square).

We call this method MulticropNCE loss, which can be expressed as:

$$\mathcal{L}_{MulticropNCE}(G, F, M, X) = \mathbb{E}_{x \sim X} \sum_{l=1}^{L} \sum_{s=1}^{S_l} \ell\left(\hat{z}_l^s, z_l^s, m_l^N\right). \quad (2)$$

*C. Domain consistency loss*

Theoretically, although the semantics and content of the generated images and the real $\mathcal{Y}$ domain images are different, they should have some similarities. They have the same style. In our set, the real $\mathcal{Y}$ domain image also extracts the features we need through $G_{encoder}$. We use two domain projection network (Hf, Hr) to extract depth style features (H represents our domain projection network, which includes a convolution layer, an average pooling layer, and a three layer linear.). Distance of these depth style features can be measured by L1 loss. By making the depth features of the domain similar, the generated image can become more realistic. Domain consistency loss is as follows:

$$\mathcal{L}_{\text{domain}}(G, Hr, Hf) = \left\| Hr(G_{encoder}(y)) - Hf(G_{encoder}(G(x))) \right\|_1^l, \quad (3)$$

where $l$ means that we use domain consistency loss for each selected layer. We guarantee the authenticity of the generated image by forcing the depth features of the generated image to be similar to the real $\mathcal{Y}$ domain image.

*D. DCA network*

In image-to-image translation, our task is mainly to focus on the style domain of images. Theoretically, the importance of each channel of features is different. We need to pay more attention to the channels related to the domain. The channel attention module aims to add weight to each channel of the feature. When the channel attention module is not used, the weight of each channel of the feature is the same. After using the attention module, the weight of each feature channel becomes different. This method makes the neural network focus on more important channels. We propose a new attention mechanism, which embeds the GAP coordinate information and the GMP coordinate information into the channel. The detail framework of DCA is shown in **Fig.4**. Before the input features are sent to DCA network, the weight of each channel is the same. After the input features pass through DCA network, the channel importance of each feature changes. Through DCA network, we can make generator pay more attention to the more related channels.

Unlike SE[19], which converts feature tensors into single dimension vectors through 2D average pool, DCA aggregates features along two spatial directions using average pooling and maximum pooling respectively. In this way, each location can capture the horizontal and vertical global information and dependencies, and we obtain GAP and GMP coordinate information. This coordinate information can increase the receptive field of network. Then, the maximum pool and average pool features are simply fused (features are added) to obtain the hybrid pool feature. The fused features contain rich information. Finally, the final weight of each channel is obtained through two layers linear for the obtained features. Our DCA network is very simple and can be flexibly plugged into various networks.

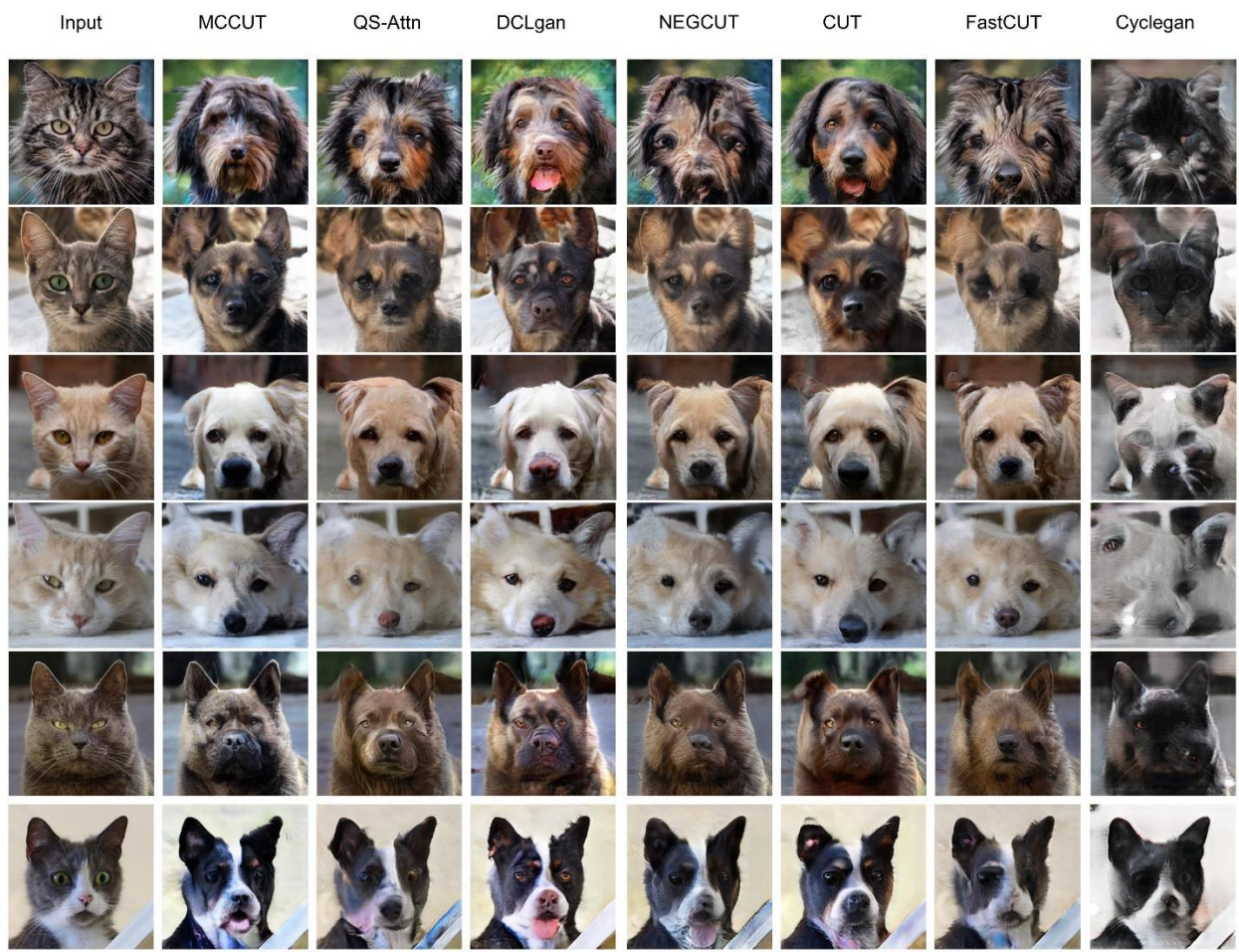

**Fig.5.** Visual comparison with all baselines on the Cat→Dog dataset. Note MCCUT, NEGCUT, DCLGAN and CycleGAN are trained for 200 epochs on the dataset. Our MCCUT can produce much clearer images with more natural details. The last row is a failure case on the Cat→Dog, our method fails to identify fur of animals sometimes.

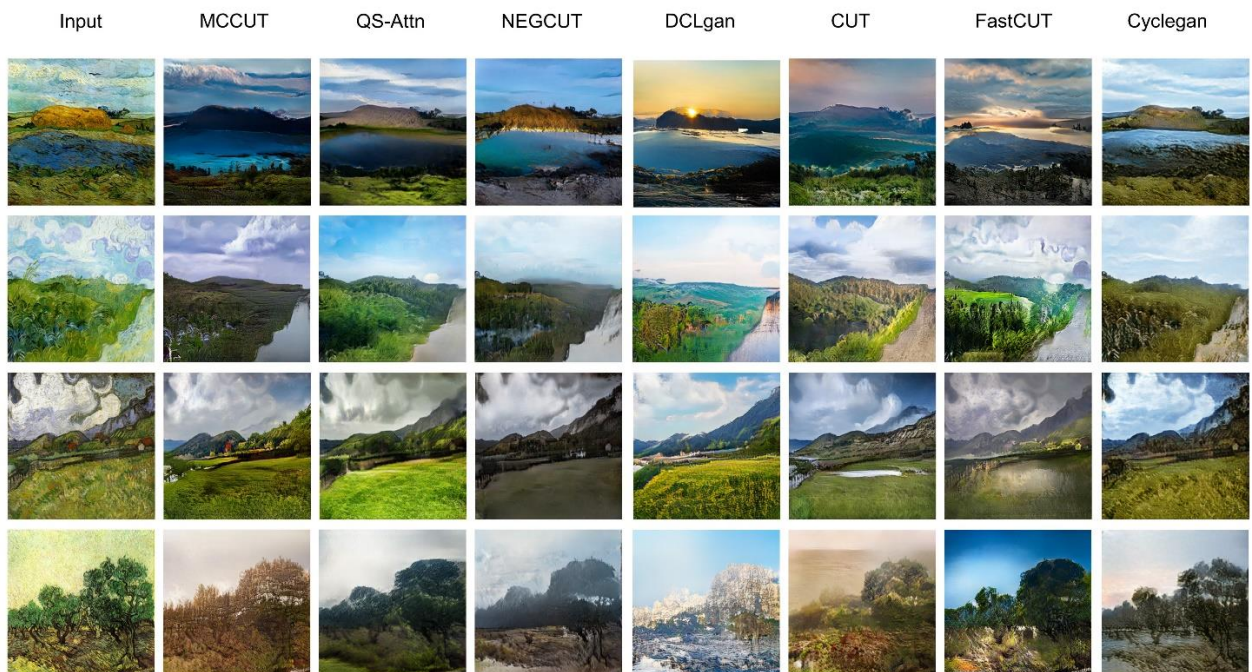

**Fig.6.** Visual comparison with all baselines on the Van Gogh→Photo dataset. Note MCCUT, NEGCUT, DCLGAN and CycleGAN are trained for 200 epochs on the dataset.

TABLE 1 QUANTITATIVE COMPARISON WITH ALL BASELINES ON THE HORSE→ZEBRA, CAT→DOG, SUMMER→WINTER, ORANGE→APPLE, VANGOGH→PHOTO DATASETS. NOTE THAT THE KID IS THE KERNEL INCEPTION DISTANCE ×100.

| Method | Horse→Zebra FID | KID | Cat→Dog FID | KID | Summer→Winter FID | KID | Orange→Apple FID | KID | Vangogh→Photo FID | KID |
|---|---|---|---|---|---|---|---|---|---|---|
| CycleGAN | 78.9 | 1.977 | 85.9 | 4.889 | 82.3 | 1.271 | 128.4 | 3.688 | 135.5 | 4.783 |
| CUT | 46.2 | **0.515** | 77.0 | **3.704** | 87.1 | 1.789 | 131.1 | 5.348 | 93.2 | 1.852 |
| DCLGAN | **46.1** | 0.659 | 75.6 | 3.766 | 91.1 | 2.002 | 125.1 | 3.749 | 93.7 | **1.743** |
| FastCUT | 73.4 | 3.212 | 95.5 | 5.509 | **78.7** | **0.747** | 132.6 | 4.124 | 105.3 | 2.598 |
| NEGCUT | 74.3 | 1.772 | 73.0 | 4.249 | 87.1 | 1.611 | 141.3 | 5.448 | 106.3 | 3.215 |
| QS-Attn | 59.3 | 1.080 | **72.3** | 3.936 | 83.2 | **1.232** | **121.9** | **3.426** | **92.2** | 1.978 |
| Ours | **36.3** | **0.385** | **61.8** | **2.663** | 80.6 | 1.295 | 119.8 | 3.152 | 91.5 | 1.768 |

*E. General Objective function*

We expect the generated image to be as similar as possible to the real target domain image, and the generator $G$ can be mapped from the $\mathcal{X}$ domain to the $\mathcal{Y}$ domain through the adversarial loss[9]. The adversarial loss[9] is as follow:

$$\mathcal{L}_{\text{GAN}}(G,D,X,Y) \lim_{x \to \infty} \\ = \mathbb{E}_{y \sim Y} \log D(y) + \mathbb{E}_{x \sim X} \log(1 - D(G(x))), \quad (4)$$

We refer to the CUT[11] settings and add an identity loss[10] to ensure that the generator not change. We take the real image in the $\mathcal{Y}$ domain as the input image, and use the MulticropNCE loss as the identity loss. Identity loss is as follow:

$$\mathcal{L}_{\text{identity}}(G,F,Y) = \mathbb{E}_{y \sim Y} \sum_{l=1}^{L} \sum_{s=1}^{S_l} \ell(\hat{z}_l^s, z_l^s, m_l^N). \quad (5)$$

Under the multi-crop contrastive learning, the corresponding patches in the input and output images should maintain similar contents. We add domain consistency loss to ensure the authenticity and quality of the generated image. We add identity loss by default. The overall objectives are as follows:

$$\mathcal{L}(G,D,F,H_f,H_r) \\ = \lambda_{GAN} \mathcal{L}_{GAN}(G,D,X,Y) \\ + \lambda_{NCE} \mathcal{L}_{MulticropNCE}(G,F,M,X). \quad (6) \\ + \lambda_{dom} \mathcal{L}_{domain}(G,F,H_f,H_r) \\ + \lambda_{ide} \mathcal{L}_{identity}(G,F,Y)$$

We set $\lambda_{GAN}=1$, $\lambda_{NCE}=1$, $\lambda_{dom}=10$, $\lambda_{ide}=1$, Following this default setting, MCCUT has achieved the best results on multiple datasets.

IV. EXPERIMENTS

Here, comprehensive experiments are conducted to evaluate the performance of MCDUT. we compare our algorithms with all baselines on different datasets. All the experiments are implemented by pytorch 1.8 on a system with an NVIDIA GeForce RTX 3080 GPU.

*A. Implementation Details*

**Datasets**

Horse → Zebra[11] contains 1067 horse images, 1344 zebra images as the training set. We use 120 horse images as the test images. MCDUT trained 400 epochs in this dataset. this dataset was collected from ImageNet[51].

Cat → Dog[52] contains 5153 images of cats and 4739 images of dogs. We used 500 images of cats as test images. MCDUT trained 200 epochs in this dataset.

Van Gogh → Photo[11] contains 400 Van Gogh paintings and 6287 photos. We used 400 Van Gogh images as test images. MCDUT trained 200 epochs in this dataset.

Orange ↔ Apple[11] contains 1019 orange images and 995 apple images. We used 248 orange images and 266 apple images as the test set. MCDUT trained 400 epochs in this dataset.

Summer → Winter[11] includes 1231 photos in summer and 962 photos in winter. We used 309 winter photos as test images. MCDUT trained 400 epochs in this dataset.

**Training details**

The implementation of MCDUT is mainly based on the CUT. Our training settings are very similar to CUT. We use Hinge GAN loss for adversarial loss. We use a generator based resnet and a discriminator based PatchGAN. Different from CUT, we use multicropNCE instead of patchNCE. Our encoder is defined as the first half of the generator. Our default setting is to extract multi-layer features from the five layers of the encoder, namely RGB pixels, the first and second down sampling convolutions, and the first and fifth resnet blocks. In features of each layer, we randomly sample 256 patches, and apply the two-layer linear network F to obtain 256 final features. Domain consistency loss also use features of these five layers. Additionally, we explored the role of DCA network in various locations of the network. we found that DCA inserted after the down sampling convolutions has the best effect. We use the Adam optimizer, $\beta 1 = 0.5$, $\beta 2 = 0.999$。 The batch size we used is 1, and all training images are loaded into $286 \times 286$, then cut to $256 \times 256$ blocks. MCDUT trains 400 epochs on each dataset, the learning rate is 0.0002 (except for individual datasets), and the learning rate

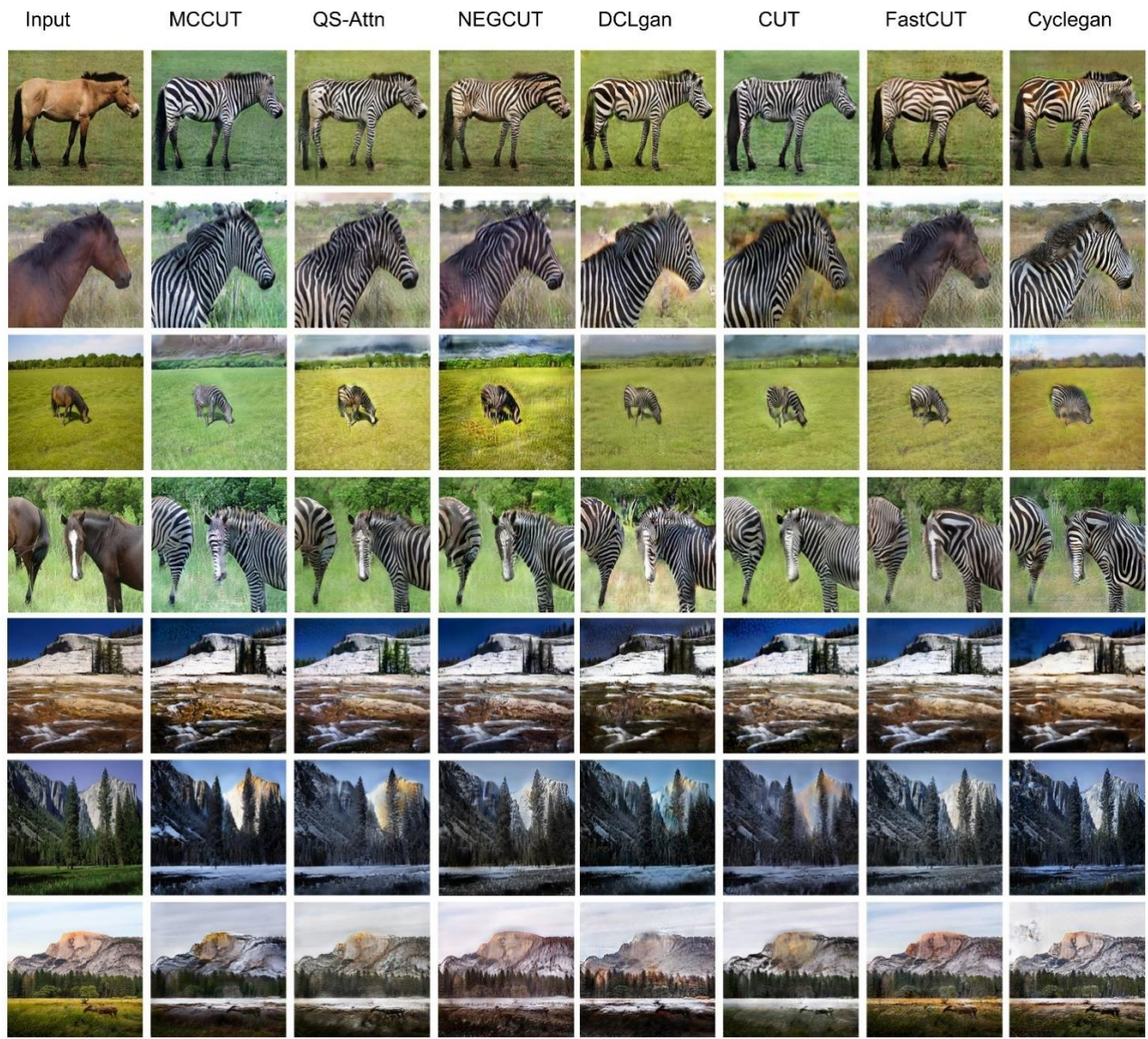

**Fig. 7.** Visual comparison with all baselines on the Horse→Zebra datasest and Summer→Winter datasest. Our MCCUT can produce more natural images. MCCUT and all the baselines are trained for 400 epochs on the two datasets.

starts to decay linearly to 0 after 200 epochs.

*B. Evaluation*

Frechet Inception Distance (FID) [53] is often used as a metric to evaluate the quality of generated images, which is specially used to evaluate the performance of GAN. FID [53] measures the similarity of two groups images from the similarity of image visual feature statistics, which is highly corresponding to human perception. A lower FID[53] means that the two sets of images are more similar, or the statistics of the generated image and the real image are more similar. The lower FID[53] is, the more realistic the generated images are. Kernel Inception Distance (KID) [54] computes the squared Maximum Mean Discrepancy between the representations of generated and real images. When the number of test images is less than the dimensionality of the inception features, KID is a more reliable metric. A lower KID means that the real and generated images are more visually similar. We mainly use FID [53] and KID [54] to estimate the quality of the generated images.

*B. Comparison*

**Table 1** shows the quantitative results of MCDUT compared with all baselines on five datasets, including Horse → Zebra, Cat → Dog, Summer → Winter, Orange → Apple, and Summer → Winter. We mainly use FID and KID ( the Kernel Inception Distance$\times 100$) scores as our quantitative metrics. For the metric of FID, the generated results of our model are more realistic than other methods on the five datasets. Obviously, our algorithm performs better than all the baselines. **Fig.5** and **Fig.6** show the visual results of random selection on Cat → Dog and Van Gogh → Photo datasets. Compared with all baselines, our MCDUT has the ability to generate the domain-relevant features accurately. MCDUT trains only 200 epochs on the two datasets.

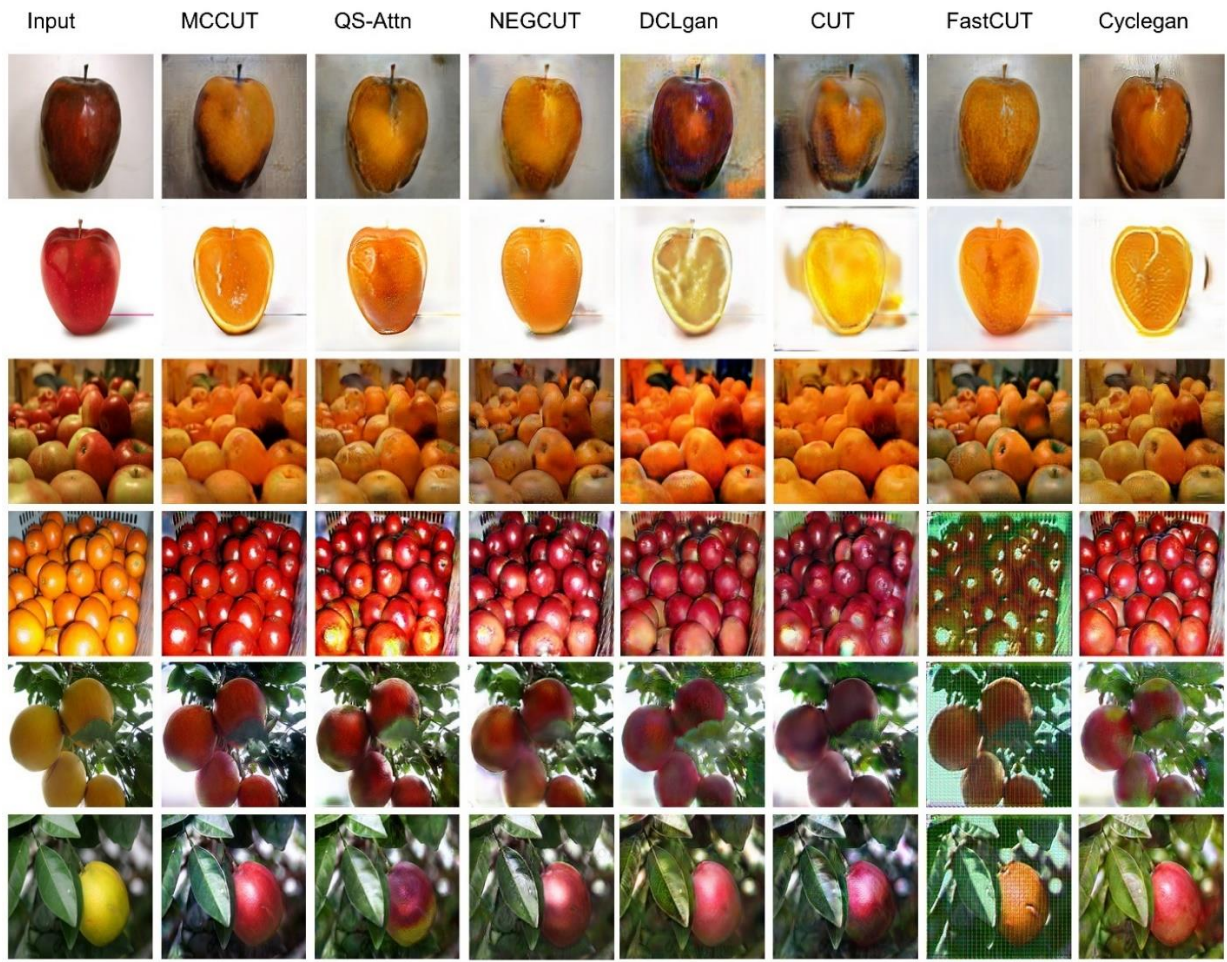

**Fig.8.** visual results comparison with all baselines on the Orange → Apple dataset. We also conduct comparison experiment about Apple → Orange. Our MCCUT can produce more real the generated images with more color details. MCCUT and all the baselines are trained for 400 epochs on the dataset.

MCDUT has achieved great success on Cat → Dog. Our MCDUT can not only generate realistic images, but also diverse results. However, there are also some examples of failure, which don't achieve ideal results in distinguishing fur of cat with exaggerated expressions. We randomly pick qualitative results on the Horse→Zebra and Summer→Winter datasets in **Fig.7**. Our MCDUT performs both geometry changes and texture changes. In the Horse→Zebra task, the MCDUT is brighter and more realistic in the structure of zebra and texture features, while other work is slightly rough. Our method is easier to distinguish between foreground and background, which proves the effectiveness of multi-crop contrastive learning. MCDUT can retain image information when processing landscape images, which makes the generated images more realistic and clear.

We randomly pick three samples on Orange → Apple and Apple → Orange in **Fig.8**. MCDUT has achieved better results in the cycle mapping of domains. Our MCDUT performs both geometry changes and texture changes. The images we generate can not only retain the features of the original image, but also make the generated images more realistic, and colorful. This proves that domain consistency loss has achieved great success. Image-to-image translation mainly depends on domain-to-domain translation. Our method has achieved the best results in many tasks, and has been greatly improved in many aspects, which proves the feasibility of the method. Our method obtains SOTA results.

*D. Ablation study*

In comparison experiments, MCDUT shows better performance than other methods. In this section, we discuss the measures that improve the performance of the model. We use the control variable method to study each of our contributions. We mainly used two tasks in ablation research, including Horse → Zebra, Cat → Dog. We conduct a series of ablation studies in the following three areas:

**The strategy of crops**: In this section, we only discuss the strategy of crops in patch-wise contrastive learning. In **Table 2**, the quantitative results on the Horse→Zebra dataset show that the strategy using one center crop and two random crops achieves the best value of FID and KID. Compared with CUT, the strategy using one center crop achieves competitive results, indicating that we can extract the key information of the input images via the center-cropping method. The strategy using one center crop and five random crops achieves the unsatisfactory results because of large amount of redundant information.

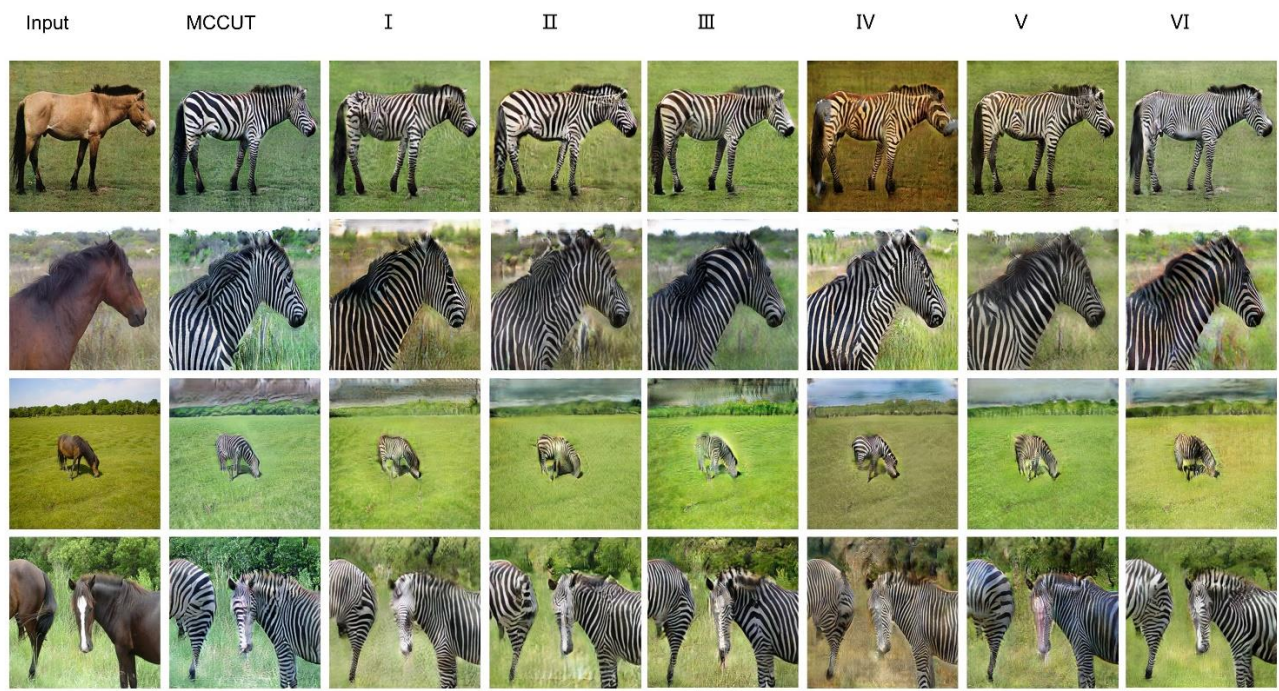

**Fig. 9.** Ablation study about each of our contribution on the Horse → Zebra dataset. The leftmost column are input images. Except for MCCUT ,the remaining columns are generated images using I-VI model.

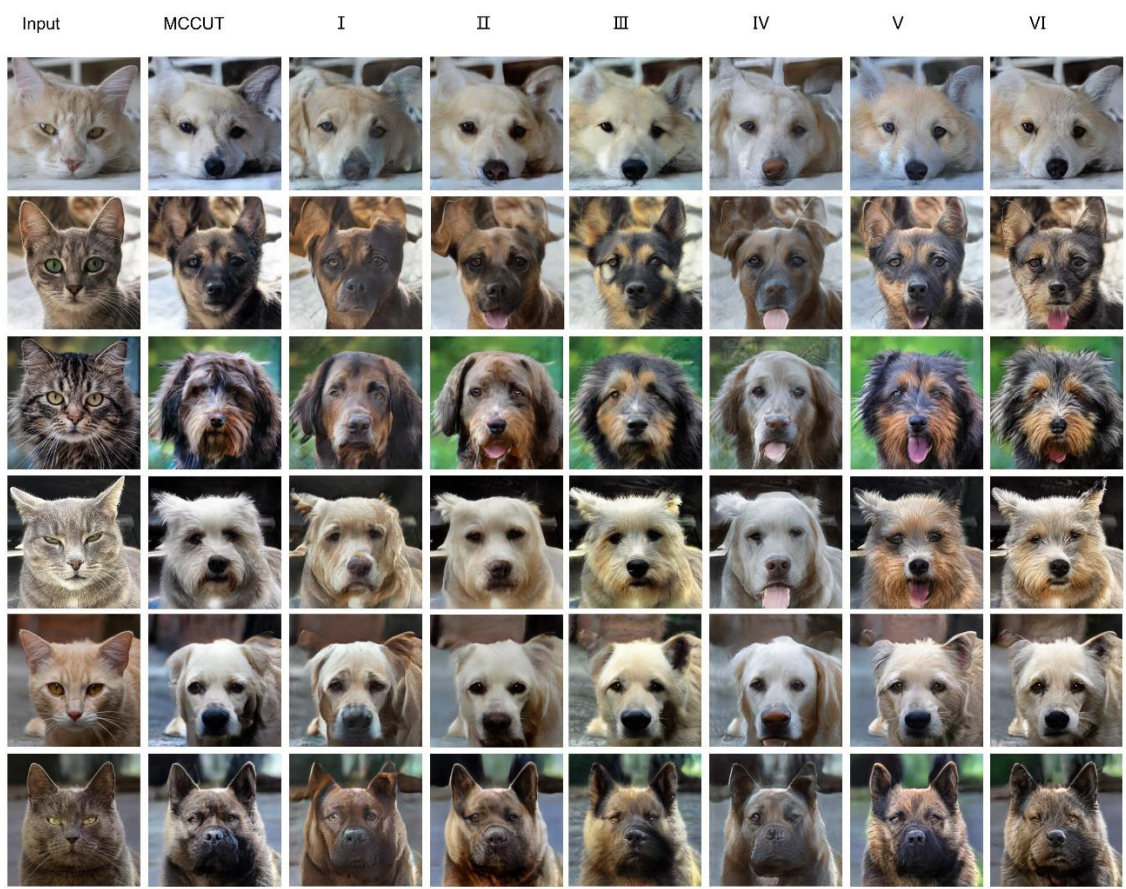

**Fig. 10.** Ablation study about each of our contribution on the Cat → Dog dataset. The leftmost column are input images. Except for MCCUT ,the remaining columns are generated images using I-VI model.

TABLE 2 QUANTITATIVE RESULTS FOR THE STRATEGY OF CROPS IN PATCH-WISE CONTRASTIVE LEARNING. NOTE THAT THE KID IS THE KERNEL INCEPTION DISTANCE ×100.

| Num_centercrop | Num_randomcrop | FID | KID |
|---|---|---|---|
| 1 | 0 | 46.9 | 0.527 |
| 2 | 0 | 51.3 | 0.879 |
| 0 | 1 | 76.8 | 2.231 |
| 1 | 1 | 43.7 | 0.483 |
| **1** | **2** | **39.1** | **0.428** |
| 1 | 3 | 42.2 | 0.467 |
| 1 | 4 | 47.7 | 0.573 |
| 1 | 5 | 53.4 | 1.132 |
| CUT | | 46.2 | 0.515 |

TABLE 3 QUANTITATIVE RESULTS FOR EFFECTIVENESS OF EACH OF OUR CONTRIBUTIONS. NOTE THAT THE KID IS THE KERNEL INCEPTION DISTANCE ×100.

| Method | Horse→Zebra | | Cat→Dog | |
|---|---|---|---|---|
| | FID | KID | FID | KID |
| I | 39.1 | 0.483 | 75.4 | 3.124 |
| II | 41.4 | 0.502 | 69.2 | 2.971 |
| III | 44.1 | 0.514 | 70.5 | 3.092 |
| IV | 42.9 | 0.497 | 74.6 | 3.143 |
| V | 39.7 | 0.488 | 68.0 | 2.784 |
| VI | 40.8 | 0.472 | 74.2 | 3.165 |
| CUT | 46.2 | 0.515 | 77.0 | 3.704 |
| MCDUT | **36.3** | **0.385** | **61.9** | **2.663** |

**Effectiveness of each of our contributions:** As shown in **Table 3**, we show the metrics results. The visual results are shown in **Fig.9** and **Fig.10**. Our base model is CUT. (I) We just replace patchNCE with multicropNCE in the base model. (II) Only domain consistency loss is added to the base model. (III) Only DCA is added to the base model. (IV) We replace patchNCE with multicropNCE, and add domain consistency loss in the base model. (V) We add domain consistency loss and DCA network to the base model. (VI) We replace patchNCE with multicropNCE and add DCA network in the base model. In **Table 3**, models I, II, and III outperform CUT, reflecting the effectiveness of my each contribution. However, compared with models I and II, the performance of model IV slightly decreased, indicating that there is a certain imbalance between the multi-cropping contrastive loss and the domain consistency loss. The metric of model V is better than models II and III, proving that the DCA is related to the domain translation, and the DCA can better assist the domain consistency loss. MCDUT outperforms all models, representing effectiveness of my all settings. MCDUT achieves the best results over the ablation experiments. Our multicropNCE is better than patchNCE and can make better use of contrastive learning. Domain consistency loss can improve the quality of the generated images. The DCA we designed may be more interested in the style of the image than the content of the image, which is appropriate for image-to-image translation tasks.

**Ablation studies of dual coordinate attention:** Firstly, we discuss effectiveness of GAP branch and GMP branch of dual coordinate attention (DCA) in MCDUT. **Table 4** shows the metrics results on the Horse→Zebra dataset. The dual setting achieves the best metrics results, proving that our setup is working. In addition, We also compare DCA network with the current popular attention modules, as shown in **Table 5**. It proves that our proposed DCA is more suitable for image-to-image translation task.

TABLE 4 QUANTITATIVE RESULTS FOR ABLATIONS OF GMP AND GAP IN DUAL COORDINATE ATTENTION. NOTE THAT THE KID IS THE KERNEL INCEPTION DISTANCE ×100.

| GMP Branch | GAP Branch | FID | KID |
|---|---|---|---|
| √ | × | 38.6 | 0.403 |
| × | √ | 37.9 | 0.394 |
| √ | √ | **36.3** | **0.385** |

TABLE 5 QUANTITATIVE RESULTS FOR ABLATIONS COMPARED WITH OTHER ATTENTION MODULES. NOTE THAT THE KID IS THE KERNEL INCEPTION DISTANCE ×100.

| Method | Horse→Zebra | | Cat→Dog | |
|---|---|---|---|---|
| | FID | KID | FID | KID |
| +SE | 38.1 | 0.392 | 66.4 | 2.891 |
| +CBAM | 39.3 | 0.413 | 69.5 | 2.933 |
| +DCA | **36.1** | **0.385** | **61.8** | **2.663** |

V. CONCLUSION

In this paper, we propose a novel contrastive learning framework for unpaired image-to-image translation, MCDUT. We obtain the negatives from the multi-crop views, which can improve the quality of negatives. We also introduce a domain consistency loss that encourages the generated images to be close to the real images in the embedding space of same domain. Moreover, we present a DCA network to improve the performance of our proposed model. In multiple datasets, MCDUT achieves the best results compared with the previous methods. Our design can better deal with various tasks of image translation. In the ablation study, we have proved the effectiveness of various settings of MCDUT. Our proposed DCA network outperforms the current attention networks in image-to-image tasks. Finally, we believe that our work will initiate further research on unsupervised image-to-image translation.

**CONFLICTS OF INTEREST**
The authors declare no conflflict of interest.

**DATA AVAILABILITY STATEMENT**
Datasets derived from public domain resources. The data that support the findings of this study are available from the corresponding author upon reasonable request.

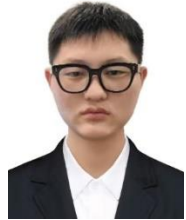
**Cheng-Wei Hu** received B.S. degree in Software Engineering from Jin Ling Institute of Technology. He is currently pursuing the M.S. degree with the Department of Computer Science & Technology, Nanjing Normal University. His reserach interests include object detection and image-to-image translation.

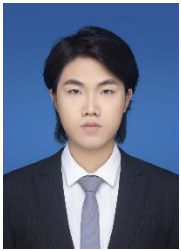
**Chen Zhao** received the B.S. degree from Zhen Zhou University of Light Industry. He is currently pursuing the M.S. degree with the Department of Computer Science & Technology, Nanjing Normal University. His research interests include contrastive learning and image generation.

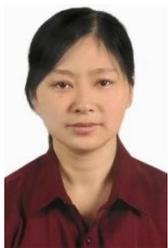
**Wei-Ling Cai** received a B.S. and Ph.D. degrees in Computer Science from Nanjing University of Aeronautics and Astronautics, China, in 2003 and 2008, respectively. At present, she is Associate Professor at the Department of Computer Science & Technology, Nanjing Normal University. Her research interests include machine learning, pattern recognition, data mining, and image processing. In these areas she has published over 20 technical papers in refereed international journals.

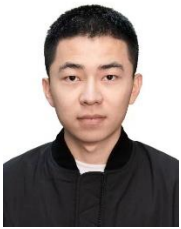
**Zheng Yuan** received B.S. degree in computer science from Henan University. He is currently pursuing the M.S. degree with the Department of Computer Science & Technology, Nanjing Normal University. His reserach interests include contrastive learning and instance segmention.